\documentclass{llncs}
\usepackage{makeidx}  
\usepackage{algorithm}
\usepackage[noend]{algorithmic}
\usepackage{qtree}
\usepackage[table]{xcolor}
\usepackage{graphicx} 

\begin{document}
\mainmatter              
\title{Intelligent Self-Repairable Web Wrappers}
\titlerunning{Intelligent Self-Repairable Web Wrappers}  
%
\author{Emilio Ferrara\inst{1} \and Robert Baumgartner\inst{2}}
\authorrunning{Emilio Ferrara and Robert Baumgartner} 
%
\tocauthor{Emilio Ferrara and Robert Baumgartner}
\institute{Dept. of Mathematics, University of Messina, V. Stagno d'Alcontres 31, 98166 Italy\\
\email{emilio.ferrara@unime.it}
\and
Lixto Software GmbH, Favoritenstrasse 16/DG, 1040 Vienna, Austria\\
\email{robert.baumgartner@lixto.com}
}

\maketitle              

\begin{abstract}
The amount of information available on the Web grows at an incredible high rate.
Systems and procedures devised to extract these data from Web sources already exist, and different approaches and techniques have been investigated during the last years.
On the one hand, reliable solutions should provide robust algorithms of Web data mining which could automatically face possible malfunctioning or failures.
On the other, in literature there is a lack of solutions about the maintenance of these systems.
Procedures that extract Web data may be strictly interconnected with the structure of the data source itself; thus, malfunctioning or acquisition of corrupted data could be caused, for example, by structural modifications of data sources brought by their owners.
Nowadays, verification of data integrity and maintenance are mostly manually managed, in order to ensure that these systems work correctly and reliably.
In this paper we propose a novel approach to create procedures able to extract data from Web sources -- the so called \emph{Web wrappers} -- which can face possible malfunctioning caused by modifications of the structure of the data source, and can automatically repair themselves.
\end{abstract}

\section{Introduction}
The actual panorama of distribution of information through the Web depicts a clear situation: there is an incredible amount of data delivered under the form of Web data sources and a corresponding need of capability of mining this information in a reliable and efficient way.
Mining information from Web sources is a task which can obviously be useful in several different area of the knowledge.
Moreover, this topic interests both the academia and the enterprises.
For example, consider the following scenarios: i) a research group which needs to acquire a dataset of information delivered through online services, say for example an online database publishing, day by day, information about the mapping of some genes; ii) a company for which it is essential, for marketing and product placement, to monitor the trends of pricing of services offered by its competitors, provided through the Web.
Both the two actors need to extract, possibly, a huge amount of data during an extend period of time (e.g., months), at regular intervals (say, each day).
One important aspect in both the cases is the reliability and the quality of data extracted.
It is utterly important that acquired information is correct, because the research group can not accept corrupted data and the comparison with competitors would fail in case of bad product data.

These two examples highlight common requirements in the panorama of Web data mining, and depict different related problems.
Although in literature some techniques to design systems for the extraction of data from Web sources have been presented, there is a lack of work in the area of their maintenance.
An ample number of questions and problems related to the possibility of automatizing the process of maintenance are still uncovered.
This work tries to focus on some aspects related to the maintenance of these systems.
We first introduce the theoretical background required to create intelligent procedures of Web data extraction.
Then, we explain how to face malfunctioning likely to happen during the extraction process, for example caused by modifications in the structure of the data source.
The second point in particular is the main focus of this work.
Let us contextualize this problem: essentially there exist two different approaches to extract information from Web sources.
The first one relies on machine learning platforms \cite{esposito2000machine}; a system analyzes, possibly, huge amount of positive and negative examples during a training period, and, then, it infers some set of rules that makes it able to perform its tasks in the same domain or Web site.
Different approaches rely on logic-based algorithms which analyze the structure of the data source and induct some procedures to extract required information exploiting structural characteristics of the Web source to identify and find required data.
The second approach utilizes the knowledge a human can bring in about a particular site or domain. The wrapper is generated in a way that the human creates the rules and navigation paths together with the system in a supervised and interactive fashion. Still, the system can assist the wrapper designer
and offer possibilities that make the wrapper execution as robust as possible, even in case of structural changes. From now, in this work we assume that the platform we are going to describe and improve adopts the latter philosophy.

\paragraph{Organization of the paper}
We describe related work in Section \ref{related-work}.
In Section \ref{comparing-trees}, the algorithmic background is introduced, describing an efficient tree matching technique.
Section \ref{wrappers} covers the design of robust and adaptable procedures of Web data extraction, henceforth called \textsl{intelligent self-repairable Web wrappers}.
Then, in Section \ref{repairing} we describe the adaptation process during wrapper execution.
We explain how these procedures can automatically, in an autonomous way, face malfunctioning, trying to adapt themselves to the modifications that possibly caused problems.
A prototype has been implemented on top of a state-of-the-art extraction platform, the Lixto Visual Developer. Performance of this  system are shown in Section \ref{performances}, by means of precision and recall scores.
Section \ref{conclusions} concludes summarizing our main achievements and depicting some future work.

\section{Background and Related Work} \label{related-work}
We split related literature in three main topics: i) Web data extraction systems; ii) maintenance and related problems; iii) tree matching algorithms.

\paragraph{Web data extraction systems}
The work related to systems of Web information extraction is manifold but well depicted by several surveys.
Laender et al. \cite{laender2002brief} provided the first rigorous taxonomical classification of Web data extraction systems.
Kushmerick \cite{kushmerick2003finite} classified several finite-state approaches to generate wrappers, such as the wrapper induction, natural language processing approaches and hidden Markov models.
Sarawagi \cite{sarawagi2008information} provided the most comprehensive survey on the information extraction panorama.
This work covers different existing techniques explaining several approaches.
In the last years, first Baumgartner et al. \cite{baumgartner2009web} and later Ferrara et al. \cite{ferrara2011web} provided two different surveys on the discipline of Web data extraction.
The first is mainly addressed to practitioners, the latter focuses on application fields of this discipline.

\paragraph{Maintenance and related problems}
Although some interesting work,
we can identify a general lack of solutions provided in the area of the Web wrapper maintenance.
Kushmerick \cite{kushmerick1999regression,kushmerick2000wrapper} for first introduced the concept of wrapper maintenance as the process of verifying the correct functioning of the data extraction procedures and manually, automatically or in a semi-automatic way, intervene in case of malfunctioning.
Lerman and Minton \cite{lerman2003wrapper}, instead, faced both the problems of verifying the correctness of data extracted by a wrapper and eventually try to repair it.
Their approach is a mix of machine learning techniques.
Another approach based on machine learning has been provided by Chidlovskii \cite{chidlovskii2003automatic}; he described a system which can automatically classify Web pages in order to extract information from those pages which can be handled adopting both conventional extraction rules and ensemble methods of machine learning, such as the content features analysis.
Meng et al. \cite{meng2003schema} developed the SG-WRAM (Schema-Guided WRApper Maintenance) slightly modifying the perspective of Web wrappers generation, observing that changes in Web pages, even substantial, always preserve syntactic features (i.e., syntactic characteristics of data items like data patterns, string lengths, etc.), hyperlinks and annotations (e.g., descriptive information representing the semantic meaning of a piece of information in its context).
Finally, another heuristic approach has been presented by Raposo et al. \cite{raposo2005automatically}; they adopted a collected sample of positive labeled examples during the normal execution of the wrappers, to be exploited in case of malfunctioning, in order to re-induct the broken wrapper ensuring a good accuracy of the process.

\paragraph{Tree Matching}
In general, the process of comparing the structure of two trees is a well-known classic problem.
The possibility of transforming a tree into another one, through a sequence of (possibly different) operations, is another well-known algorithmic challenge, namely the \emph{tree editing} problem.
The minimum number of elementary transformations, such as adding/removing nodes, relabeling nodes or moving nodes, represents the \emph{distance} between two trees.
This value can be used to represent the measure of dissimilarity between two trees.
The tree edit distance problem is a well-known NP-hard problem \cite{bille2005survey}.
Several approximate solutions have been advanced during the years; the most appropriate algorithm to face the problem of matching up similar trees, has been suggested by Selkow \cite{selkow1977tree}.
This technique relies on the concept of finding isomorphic elements present in both the two compared trees, implementing a light-weight recursive top-down resolution during which the algorithm evaluates the position of nodes to measure the degree of isomorphism between them, analyzing and comparing their sub-trees.
Different versions of this algorithm exist; each of them presents some optimizations.
Ferrara and Baumgartner \cite{ferrara2010automatic,ferrara2011design} so as Yang \cite{yang1991identifying} adopt \emph{weights}, obtaining a variant of this algorithm with the capability of discovering clusters of similar sub-trees.
An interesting evaluation of the simple tree matching and its weighted version, presented by Kim et al. \cite{kim2008web}, has been performed exploiting these two algorithms to extract information from HTML Web pages.
These optimized algorithms underly the design of our self-repairable Web wrappers.

\section{The Tree Matching Algorithm} \label{comparing-trees}
This work relies on some assumptions: i) Web pages are represented by using DOM trees, as the HTML standard imposes\footnote{http://www.w3.org/TR/DOM-Level-2-HTML/html.html}; ii) it is possible to identify elements within a DOM tree by using the XPath language\footnote{http://www.w3.org/TR/xpath/}; iii) the logics of XPath underly the functioning of Web wrappers (this is further explained in following sections and in \cite{baumgartner2009web,baumgartner2009scalable}).
Given these milestones, the main idea of our approach is to compare two trees, one representing the original Web page and another representing the page after that some modifications occurred.
This is practical in order to automatize the adaptive process of automatic repairing of our wrappers.
To do so, we utilize a variant of the seminal Simple Tree Matching (STM) \cite{selkow1977tree}, optimized by Ferrara and Baumgartner \cite{ferrara2010automatic,ferrara2011design}.
Let \emph{d(n)} be the degree of a node \emph{n} (i.e., the number of first-level children); let $T(i)$ be the i-\emph{th} sub-tree of the tree rooted at node $T$; let $t(n)$ be the number of total siblings of a node \emph{n} including itself.
The \emph{Weighted Tree Matching} here described (see Algorithm \ref{alg:clustered}) optimizes the simple tree matching, for our specific domain.

\begin{algorithm}
\caption{WeightedTreeMatching($T^{'}$, $T^{''}$)}
\label{alg:clustered}
\begin{algorithmic}[1]
    \IF{$T^{'}$ has the same label of $T^{''}$}
        \STATE $m \leftarrow$ $d(T^{'})$
        \STATE $n \leftarrow$ $d(T^{''})$
        \FOR{$i = 0$ to $m$}
            \STATE $M[i][0] \leftarrow 0$;
        \ENDFOR
        \FOR{$j = 0$ to $n$}
            \STATE $M[0][j] \leftarrow 0$;
        \ENDFOR
        \FORALL{$i$ such that $1\leq i\leq m$}
            \FORALL{$j$ such that $1\leq j \leq n$}
                \STATE $M[i][j] \leftarrow$ Max($M[i][j-1]$, $M[i-1][j]$, $M[i-1][j-1] + W[i][j]$) where $W[i][j]$ = WeightedTreeMatching($T^{'}(i-1)$, $T^{''}(j-1)$)
            \ENDFOR
        \ENDFOR

        \IF{$m > 0$ AND $n > 0$}
            \STATE return M[m][n] * 1 / Max($t(T^{'})$, $t(T^{''})$)
            \ELSE
            \STATE return M[m][n] + 1 / Max($t(T^{'})$, $t(T^{''})$)
            \ENDIF
    \ELSE
        \STATE return 0
    \ENDIF
\end{algorithmic}
\end{algorithm}

\section{Web Wrappers} \label{wrappers}

In supervised and interactive wrapper generation, the application designer is in charge of deciding how to characterize Web objects that are used for traversing the Web and for extracting information.
It is one of the most important aspects of a wrapper to be resilient against changes (both changes over time and variations of similarly structured pages), and parts of the robustness of a data extractor depend on how the application designer configures it. 
However, it is crucial that the wrapper generation system assists the wrapper designer and suggests how to make the identification of Web objects and trails through Web sites as stable as possible. 

\subsection{Robust XPath Generation and Fall-back Strategies}

In Lixto Visual Developer (VD) \cite{baumgartner2009scalable}, a number of mechanisms are offered to create a resilient wrapper.
During recording, one task is to generate a robust XPath or regular expression, interactively and supported by the system. 
During wrapper generation, in many cases only one labeled example object  is available, especially in automatically recorded deep Web navigation sequences. 
In such cases, efficient heuristics in XPath generation and fallback strategies during replay, are required. 
Typical heuristics during recording for reliably identifying such single Web objects include:

\begin{itemize}
	\item Generalization of a chosen XPath by using form properties, element properties, textual properties and formatting properties. 
During replay, these ingredients are used as input for an algorithm that checks in which constellation to best apply this property information to satisfy the integrity constraints imposed on a rule (e.g., as result a single instance is required).
	\item DOM Structural Generalization -- starting from the full path, several generalized paths are created, using only characteristic elements and characteristic element sequences. 
A number of stable anchor points are identified and stored, from which relative paths to this object are created. 
Typical stable anchor points are identified automatically and include, e.g., the outermost table structure and the main content area (being chosen upon factors such as the longest content).
	\item Positional information is considered if the structurally generalized paths identify more than one element. 
In this case, during execution, variations of the XPath generated with this ``index heuristics'' are applied on the active Web page, removing indexes until the integrity constraints of the current rule are satisfied.
	\item Attributes and properties of elements are taken into account, in particular of the element of choice, but we also consider ancestor attributes if the element attributes are not sufficient.
	\item Attributes that make an element unique are preferred, i.e., similar elements are checked for distinguishing criteria.
	\item Attribute Values are considered, if attribute names are not sufficient. 
Attribute Value Fragments are considered, if attribute values are not sufficient (using regular expressions).
	\item The ID attributes are used as far as possible. 
If an ID is unique and meaningful for characterizing an element it is considered in the fallback strategies with a high weight. 
	\item Textual information and label information is used, only if explicitly turned on (since this might fail in case of a language switch).
\end{itemize}

\begin{figure}[!t]%
    \includegraphics[width=\columnwidth]{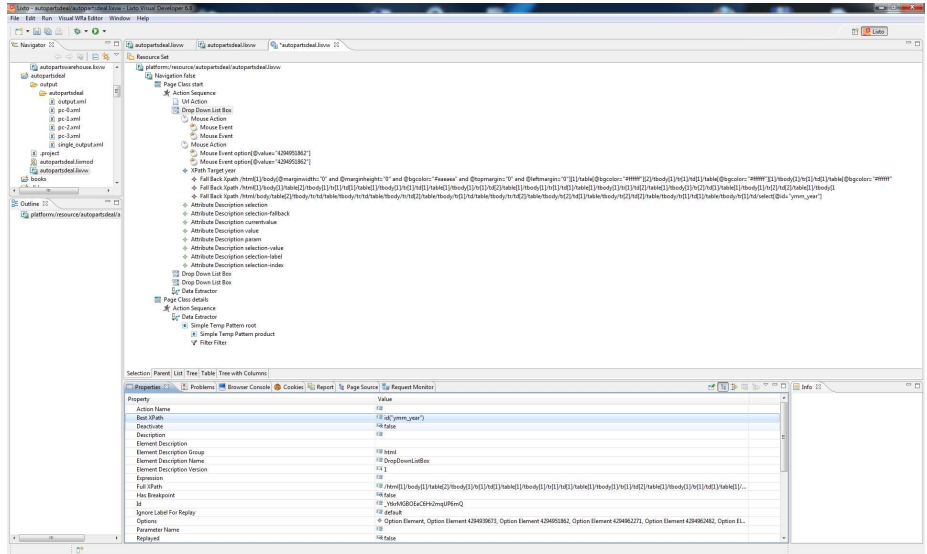}%
    \caption{Robust Web Object Detection in Visual Developer.}%
    \label{fallback}%
\end{figure}

The output of the heuristic step is a ``best XPath'' shown to the wrapper designer, and a set of XPath expressions and priorities regarding when to use which fallback strategy, stored in the configuration. 
Figure \ref{fallback} illustrates which information is stored by the system during recording. 
In this case, a drop down was selected by the application designer, and the system decided that the ``id'' attribute is the most reliable one and is chosen as best XPath. 
If this evaluation fails, the system will apply heuristics based on the (in this example, three) stored fallback XPaths, which mainly exploit form and index properties. 
In case one of the heuristics generates results that do not invalidate the defined integrity constraints, these Web objects are considered as result.

During generation of rules (e.g., ``extract'') and actions (e.g., ``click''), the wrapper designer imposes constraints on the results to be obtained, such as:

\begin{itemize}
	\item Cardinality Constraints: restrictions on the number of results, e.g., exactly one element or at least one element must be matched.
	\item Data Type Constraints: restrictions on the data type of a result, e.g., a result must be of type integer or match a particular regular expression.
\end{itemize}

Constraints can be defined individually per rule and action, or defined globally by using a schema on the output data model.

\subsection{Configuring Adaptable Wrappers}
The procedures described in the previous section do not adapt the wrapper, but address situations in which the initially chosen XPath does no longer match and simply try different ones based on this one. 
In the configuration of wrapper adaptation, we go one step beyond: on the one hand we exploit tree and string similarity techniques to find the most similar Web object(s) on the new page, and on the other hand, in case the adaptation is triggered, the wrapper is changed on the fly using the new configuration created by the adaptation algorithms.

As before, integrity constraints can be imposed on extraction and navigation rules.
Moreover, the application designer can choose whether to use wrapper adaptation on a particular rule in case the constraints are violated during runtime. 
When adaptation is chosen, alternatively to using XPath-based means to identify Web objects we store the actual result subtree. 
In case of HTML leaf elements, which are usually the elements under consideration for navigation actions, we instead store the tree rooted at the \emph{n-th} ancestor of the element, and the additional fact where the result element is located within this tree. 
In this way, tree matching can also be exploited for HTML leaf elements.

\begin{figure}[!t]%
    \includegraphics[width=\columnwidth]{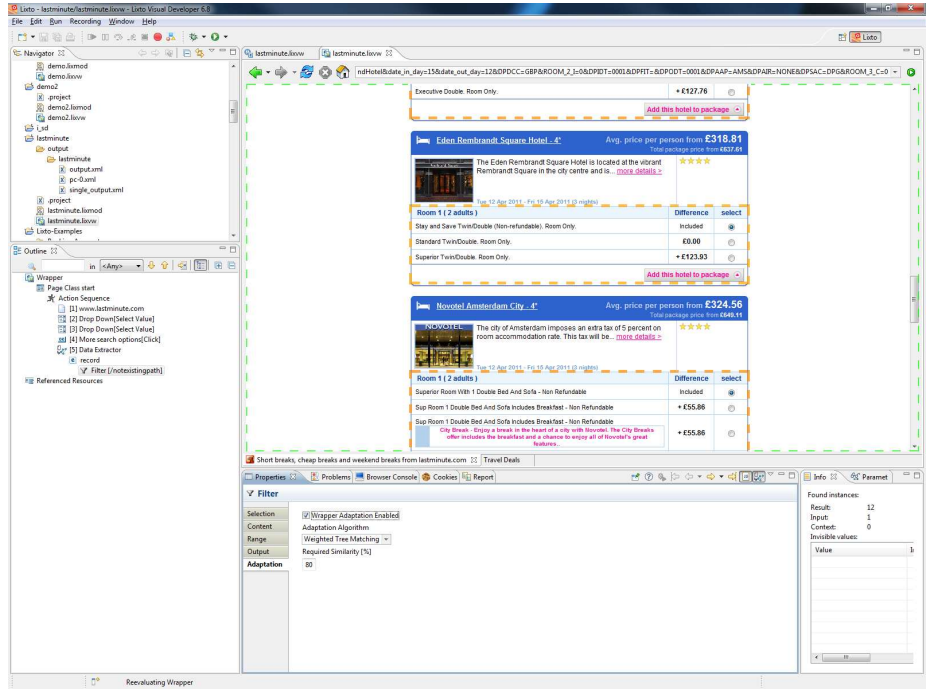}%
    \caption{Configuration of Wrapper Adaptation in Lixto VD.}%
    \label{vd1}%
\end{figure}

Wrapper designers can choose between various similarity measures: this includes in particular the Simple Tree Matching algorithm \cite{selkow1977tree} and the Weighted Tree Matching algorithm described in Section \ref{comparing-trees}. 
In future, further algorithms will extend the capabilities of the tool, e.g., a bigram-based tree matching that is capable to deal with node permutations in a more favorable fashion. 
In addition to the similarity function, one can choose certain parameters, e.g., whether to use the HTML element name as node label or instead to use spelling attributes such as \emph{class} and \emph{id} attributes. 
Figure \ref{vd1} illustrates the configuration of wrapper adaptation in Visual Developer.

\section{Automatic Wrapper Adaptation} \label{repairing}

\subsection{Self-repairing rules}

\begin{figure}[!t]%
    \includegraphics[width=\columnwidth]{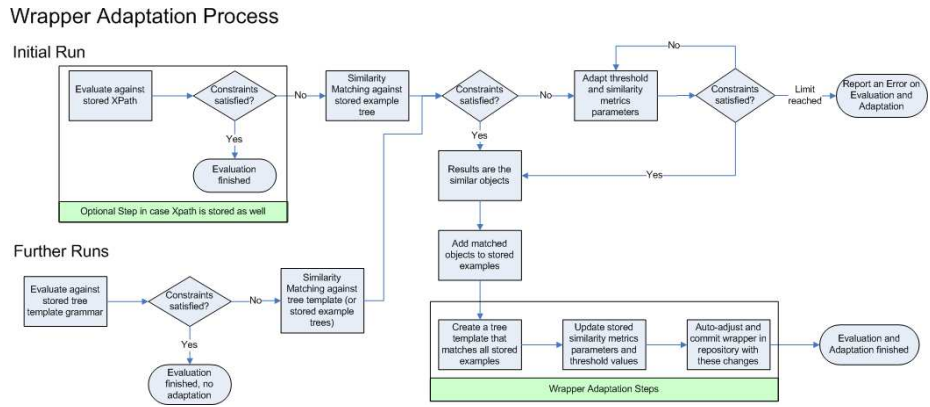}%
    \caption{Wrapper Adaptation Process.}%
    \label{ap}%
\end{figure}

Figure \ref{ap} describes the adaptation process. 
The wrapper adaptation process is triggered upon violation of defined constraints. 
In case in the initial wrapper an element is detected with an XPath, the adaptation procedure substitutes this by storing the subtree of a matched element. 
In case the wrapper definition already stores the example tree, and the similarity computation returns results that violate the defined constraints, the threshold is lowered or raised until a perfect match is generated.

During runtime, the stored tree is compared to the elements on the new page, and the best fitting element(s) are considered as extraction results. 
During configuration, wrapper designers can choose an algorithm (such as the Weighted Tree Matching), and a similarity threshold. 
The similarity threshold can be constant, or defined to be within an interval of acceptable thresholds.
During execution, various thresholds within the allowed range are considered, and the one generating the best fit with respect to the defined constraints is chosen.

As a next step, the stored tree is refined and generalized so that it maximizes the matching value for both the original subtree and the new trees, reflecting the changes of a Web page over time.
This generalization process generates a simple tree grammar, a ``tree template'' that is allowed to use occurrence indicators (one or more element, at least one element, etc.) and optional depth levels. 
In further runs, the tree template is compared against the sub trees of an active Web page during execution. 
First, the algorithm checks which trees on the new page satisfy the tree template. 
In case the results are within the defined integrity constraints, no further action is taken. 
In case the results are not satisfying, the system searches for most similar trees based on the defined distance metrics; in this case, the wrapper is auto-adapted, the tree template is
further refined and the threshold or threshold interval is automatically re-adjusted. 
At the very end of the process, the corrected wrapper is stored in the wrapper repository and committed to a versioning system to keep track of all changes.

\subsection{Wrapper Re-Induction}
In practice, single adaptation steps of rules and actions are embedded into the whole execution process of a wrapper and the adapted wrapper is stored in the repository after all adaptation steps have been concluded. 
The need for adapting a particular rule  influences the further execution steps. 

\begin{figure}[!t]%
    \includegraphics[width=\columnwidth]{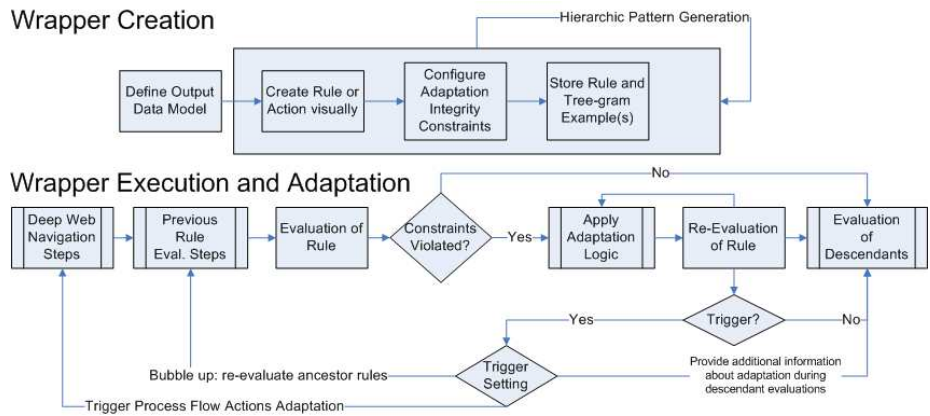}%
    \caption{Diagram of the Web wrapper creation, execution and maintenance flow.}%
    \label{flow-diagram}%
\end{figure}

Usually, wrapper generation in VD is a hierarchical top-down process -- e.g., first, a ``hotel record'' is characterized, and inside the hotel record, entities such as ``rating'' and ``room types''. 
To define a rule to match such entities, the wrapper designer visually selects an example and together with system suggestions generalizes the rule configuration until the desired instances are matched. 
To support the automatic adaptation process during runtime, as described above, the wrapper designer further specifies what it means that extraction failed. 
In general, this means wrong or missing data, and with integrity constraints one can give indications how correct results look like. 
The upper half of  Figure \ref{flow-diagram} summarizes the wrapper generation.

During wrapper creation, the application designer provides a number of configuration settings to this process. 
This includes:

\begin{itemize}
	\item Threshold Values.
	\item Priorities/Order of Adaptation Algorithms used.
	\item Flags of the chosen algorithm (e.g., using HTML element name as node label, using id/class attributes as node labels, etc.).
	\item Triggers for bottom-up, top-down and process flow adaptation bubbling.
	\item Whether stored tree-grams and XPath statements are updated based on adaptation results to be additionally used as inputs in future adaptation procedures (reflecting and addressing regular slight changes of a Web page over time).
\end{itemize}

Triggers in Adaptation Settings can be used to force adaptation of further fragments of the wrapper as depicted in the lower half of Figure \ref{flow-diagram}.

\begin{itemize}
	\item Top-down: forcing adaptation of all/some descendant rules (e.g., adapt the ``price'' rule as well to identify prices within a record if the ``record'' rule was adapted).
	\item Bottom-up: forcing adaptation of a parent rule in case adaptation of a particular rule was not successful.
	Experimental evaluation pointed out that in such cases it is often the problem that the parent rule already provides wrong or missing results (even if matched by the integrity constraints) and has to be adapted first.
	\item Process flow: it might happen that particular rule matches can no longer detected because the wrapper evaluates on the wrong page. 
	Hence, there is the need to use variations in the deep web navigation actions. 
	In particular, a simple approach explored at this time is to use a switch window or back step action to check if the previous window or another tab/popup provides the required information.
\end{itemize}

\section{Performances Measurement} \label{performances}
For our initial performance evaluation we tested the robustness of our Wrappers against real world use-cases.
Actual areas of interest for Web data extraction problems include social networks, retail market and Web communities.
We defined a total of 7 scenarios and designed 10 adaptive wrappers each.
Results, by means of precision, recall and F1-score, are as shown in Table \ref{tab-res}.
Column \emph{thresh.} represents the fixed threshold value; \emph{tp}, \emph{fp} and \emph{fn} summarize true and false positive, and false negative, respectively.
Performance obtained by using \emph{simple} and \emph{weighted} tree matching are good; these algorithms are definitely viable solutions to our initial purpose and provide high degree of reliability (F-Measure $>$ 90\%). 

\begin{table}[!t]
    \centering
    
    \begin{tabular}{|@{}c@{}  c@{}|@{}c  c  c@{}|@{}c  c  c@{}|}
			\cline{3-8}
        \multicolumn{2}{r|}{} & \multicolumn{3}{c|}{Simple Tree Matching} & \multicolumn{3}{c|}{Weighted Tree Matching} \\
    \cline{3-8}
        \multicolumn{2}{r|}{} & \multicolumn{3}{c|}{Precision/Recall} & \multicolumn{3}{c|}{Precision/Recall}\\
    \cline{3-8}

        \noalign{\smallskip}
    \hline
        Scenario & thresh. & tp & fp & fn & tp & fp & fn \\
    \hline
        Delicious & 40\% & 100 & 4 & - & 100 & - & - \\
        Ebay & 85\% & 200 & 12 & - &  196 & -  & 4 \\
        Facebook & 65\% & 240& 72 & - &  240&12 & - \\
                Google news & 90\% & 604 & - & 52 &  644 & - & 12\\
        Google.com & 80\% & 100 & - & 60 &  136 & - & 24 \\
        Kelkoo & 40\% & 60 & 4 & - & 58 & - & 2 \\
        Techcrunch & 85\% &  52 & - & 28 &  80 & - & - \\
    \hline
        Total  & - & 1356 & 92 & 140 & 1454 & 12 & 42\\
    \hline
    \hline
        Recall  & - & \multicolumn{3}{c|}{90.64\%} & \multicolumn{3}{c|}{97.19\%}\\
        Precision  & - & \multicolumn{3}{c|}{93.65\%} & \multicolumn{3}{c|}{99.18\%}\\
        F-Measure  & - & \multicolumn{3}{c|}{92.13\%} & \multicolumn{3}{c|}{98.18\%}\\
        \hline

    \end{tabular}

    \caption{Experimental performance evaluation in real world scenarios.}
    \label{tab-res}

\end{table}

\section{Conclusions and Future Work} \label{conclusions}
In literature, several implementations of systems to extract data from Web sources have been presented, but there is a lack of solutions about their maintenance.
This paper tries to address this problem, describing adaptive techniques to make Web data extraction systems, based on wrappers, self-maintainable, adopting algorithms optimized to this purpose.
So, enhanced Web wrappers become able to recognize structural modifications of Web sources and to adapt their functioning accordingly.
Characteristics of our self-repairable solution are discussed in details, providing first experimental results to evaluate its robustness.
More experimentation has to come in the next future.

Moreover, as for future work, additional algorithms would be included in order to improve the capabilities of the adaptation feature; in particular, a viable idea could be to generalize a bigram-based tree matching algorithm capable of dealing with node permutations in a more efficient way with respect to Simple Tree Matching based algorithms adopted as to date.
Similarly, the Jaro-Winkler distance could be adapted to our tree matching problem in order to better reflect missing or added node levels, so as improving performance of our adaptation process.
Finally, the tree-grammar could be extended to classify different topologies of templates (those frequently adopted by Web pages), in order to define several standard protocols of automatic adaptation, to be adopted in specific contexts.


\bibliographystyle{splncs03}
\bibliography{aixia-bib}

\end{document}